\documentclass{article}

\usepackage{arxiv}

\usepackage[utf8]{inputenc} 
\usepackage[T1]{fontenc}    
\usepackage{hyperref}       
\usepackage{url}            
\usepackage{booktabs}       
\usepackage{amsfonts,amsmath,amssymb}       
\usepackage{nicefrac}       
\usepackage{microtype}      
\usepackage{graphicx}

\usepackage{color, colortbl}

\usepackage{mathptmx,amsthm}
\usepackage{xcolor}
\usepackage{multirow}

\newcommand{\ie}{\textit{i.e.,}}
\newcommand{\eg}{\textit{e.g.,}}
\newcommand{\commenttxt}[1]{}
\newcommand{\mysubheading}[1]{\vspace{0.25cm}\noindent\textbf{#1:}}
\newcommand{\mybar}{\kern1pt\rule[-\dp\strutbox]{.8pt}{\baselineskip}\kern1pt}

\newcommand{\sntr}{\mathbf{S}_{\text{N}}}
\newcommand{\sblt}{\mathbf{S}_{\text{B}}}
\newcommand{\cmod}{\mathbf{C}_{\text{CL}}}
\newcommand{\expt}[1]{E\textsubscript{#1}}

\renewcommand{\mytagline}{} 

\theoremstyle{definition}

\title{CCESAR: Coastline Classification-Extraction from SAR Images Using CNN-U-Net Combination}
\author{Vidhu~Arora, Shreyan~Gupta, Ananthakrishna~Kudupu, Aditya~Priyadarshi, \\Aswathi~Mundayatt and~Jaya~Sreevalsan-Nair\thanks{Corresponding author; \texttt{jnair@iiitb.ac.in} \\ V. Arora, S. Gupta, A. Kudupu, and A. Priyadarshi contributed equally in developing the deep learning models.} \\
  \\
  Graphics-Visualization-Computing Lab,\\
  International Institute of Information Technology Bangalore, Karnataka 560100, India \\
  \texttt{http://www.iiitb.ac.in/gvcl} 
}

\begin{document}
\maketitle

\begin{abstract}
  In this article, we improve the deep learning solution for coastline extraction from Synthetic Aperture Radar (SAR) images by proposing a two-stage model involving image classification followed by segmentation. We hypothesize that a single segmentation model usually used for coastline detection is insufficient to characterize different coastline types. We demonstrate that the need for a two-stage workflow prevails through different compression levels of these images. Our results from experiments using a combination of CNN and U-Net models on Sentinel-1 images show that the two-stage workflow, coastline classification-extraction from SAR images (CCESAR) outperforms a single U-Net segmentation model.
\end{abstract}


\section{Introduction}
Monitoring coastline changes is a critical step in evaluating environmental changes, especially with respect to global warming and melting icecaps~\cite{tsokas2022sar}. Hence, coastline detection and extraction from remote sensing data is an important problem to solve. Synthetic Aperture Radar (SAR) is a remote sensing technology that has promise in this activity because it can penetrate through cloud cover, thus making data available under all weather conditions~\cite{tsokas2022sar, ciecholewski2024review}. At the same time, coastline extraction from SAR images is relatively new compared to optical images and is an important problem to solve~\cite{ciecholewski2024review}.

Both coastline and shoreline are defined as a physical boundary between land and water and are interchangeably used~\cite{ciecholewski2024review}. However, coastline refers to large-scale regions with changes over longer periods of time unlike shoreline, which is usually marked by the high-tide mark~\cite{oertel2005coasts}. Despite this epistemological difference between coastline and shoreline, \textit{shoreline detection} and \textit{coastline} detection are interchangeably used in literature. Deep learning models, such as U-Net, have been used effectively for land-water differentiation for coastline/shoreline~\cite{ciecholewski2024review}. Coastline detection is often solved as a classification problem where pixels are classified into binary classes, \ie~land and water.

However, the coastline studies are most often localized, and far less generalized~\cite{heidler2021hed, chang2022u}. The localized studies show good performance of deep learning models owing to the strength of the models. But the credit also goes to the fine-tuned local parameters, stemming from the low variation in the coastline properties. In this work, we hypothesize that for a generalized solution for coastline detection, it is imperative to group the coastlines based on their salient properties and compute group-based deep learning models for their detection and subsequent extraction. The rationale is that there are known perceptible differences between \textit{non-urbanized} and \textit{urbanized} coastlines, where the latter includes port infrastructure~\cite{ciecholewski2024review}. There is also a gap in a labeled dataset containing a mix of such coastline images.

We thus propose a classification step before coastline detection and extraction. In this work, as a proof of concept, we identified two classes of coastline that have \textit{perceptual} differences, and we refer to them as \textit{natural} and \textit{built} coastlines. Our proposed workflow, Coastline Classification and Extraction from SAR images (CCESAR) uses a two-stage network for classification and segmentation, which sequentially runs a Convolutional Neural Network (CNN) and a U-Net model for classification and detection of coastlines, respectively. CNN model is known to classify images and U-Net for segmentation~\cite{hossain2023automated}, thus making it an ideal choice for coastline classification in our work. Similar work is seen where segmentation using U-Net followed by classification using CNN for breast tumor segmentation~\cite{hossain2023automated}.

SAR imagery is available in several image formats and compression levels~\cite{ciecholewski2024review}. Here, we also evaluate the effectiveness of our proposed workflows for GeoTIFF images in 8-bit and 32-bit formats. The 32-bit image format includes the actual radar backscatter in float format, whereas the 8-bit image format has the same in compressed or quantized format, implying loss of information. Overall, our contributions are as follows:
\begin{itemize}
\item We propose a two-stage classification-segmentation deep learning solution, \ie~an integrated classification-segmentation workflow, CCESAR for coastline extraction from SAR images.
\item We compare the performance of CCESAR on image formats with different precision, namely, 8-bit and 32-bit GeoTIFF images.
\item As an outcome of developing deep learning models, we publish a dataset of SAR images from Sentinel-1 with mixed coastline types.
\end{itemize}

\begin{figure}
  \centering
  \includegraphics[width=0.5\linewidth]{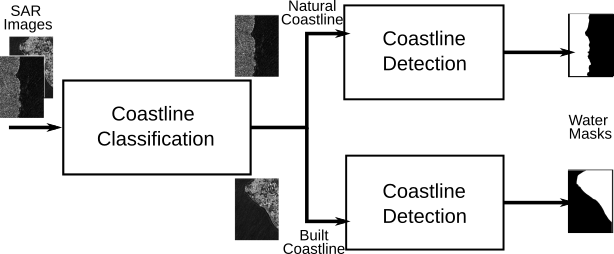}
  \caption{Two stages in our proposed model in CCESAR.}
  \label{fig:workflow}
\end{figure}

\section{Dataset Preparation}
\subsection{Data Collection}
A dataset of Sentinel-1 SAR images, downloaded on July 03, 2024, is used for this analysis. Sentinel-1 imagery was selected for its ability to distinguish between land and water through backscatter variation and its capacity to penetrate cloud cover, making it particularly useful for coastal regions~\cite{tsokas2022sar, ciecholewski2024review}. The data was sourced using the Copernicus Data Browser, \url{https://browser.dataspace.copernicus.eu/}.

To ensure that the dataset is representative of various global coastlines, random sampling was applied to clip areas of 100--150 km\textsuperscript{2} from different regions in the world, ensuring that the dataset contains a variety of coastline types and this enhances the robustness of our learning models. The regions covered included the Netherlands, London, Ireland, Spain, France, Lisbon, the USA, India, and Africa. We used datasets of GeoTIFF image format with different precision, namely, 8-bit and 32-bit. For the 32-bit dataset, Italy was additionally included in the built regions.

The imagery, captured using the Interferometric Wide Swath (IW) mode of Sentinel-1, has a spatial resolution of $10m\times 10m$. The IW mode provides higher wide-area coverage even at high resolution. We used dual-polarization settings, \ie~the VV (Vertical transmit and Vertical receive) and the VH (Vertical transmit and Horizontal receive) modes. These polarization configurations help to distinguish different surface types, making them particularly effective for coastline detection. Details of our dataset are given in Table~\ref{tab:datasets}.

\begin{table}[t]
\centering
\caption{The datasets used in CCESAR evaluation}
\label{tab:datasets}
\begin{tabular}{|c|c|c|c|}
\hline
\textbf{Dataset} & \textbf{Area Type} & \textbf{Training Count} & \textbf{Testing Count} \\ \hline
\multirow{2}{*}{8-bit}
& Natural & 200 & 40 \\ \cline{2-4}
& Built   & 209 & 40 \\ \hline
\multirow{2}{*}{32-bit}
& Natural & 201 & 40 \\ \cline{2-4}
& Built   & 199 & 40 \\ \hline
\end{tabular}

\textit{Note: All images cover an average of 100-150 km\textsuperscript{2}.}
\end{table}

\section{CCESAR: Our Proposed Workflow}
\begin{figure*}
\centering 
\includegraphics[width=\linewidth]{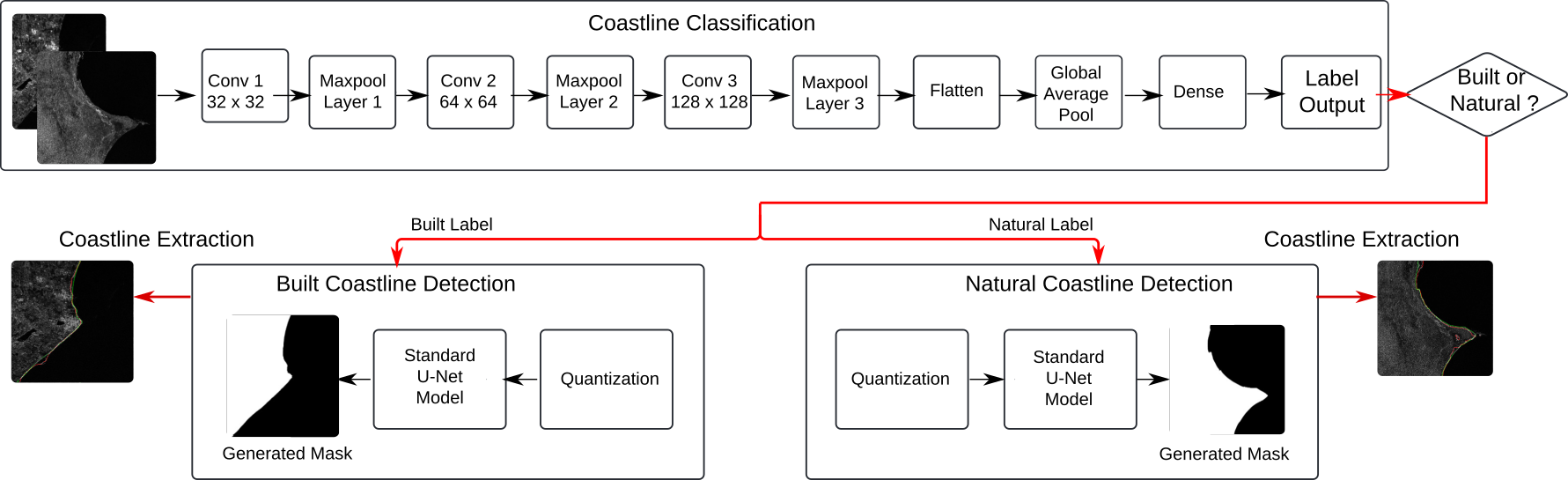}
\caption{Our proposed two-stage deep learning model for image classification and coastline segmentation in CCESAR.}
\label{fig:model}
\end{figure*}

We solve the problem of coastline extraction of SAR images consisting of natural and built coastlines using a two-stage approach: a convolutional neural network (CNN) for classification and a U-Net architecture for segmentation. These models are specifically designed to process the spatial hierarchies present in SAR imagery, ensuring both high-level feature extraction and pixel-level segmentation accuracy. Our proposed method is shown in Figure~\ref{fig:workflow}.

\subsection{Preprocessing of SAR Images}
With our curated dataset, we perform additional preprocessing steps only for 32-bit images to prepare them for high-precision analysis. On the other hand, the 8-bit GeoTIFF images were directly downloaded from the Copernicus Data Browser. The preprocessing steps of 32-bit images are:
\begin{enumerate}
    \item \textbf{Speckle Noise Reduction:} We reduce speckle noise in the input SAR images using a Lee filter with a 5x5 window size. This step helps to preserve important features, such as coastline edges, while minimizing noise.
    
    \item \textbf{Geometric Correction:} Geometric distortions in the SAR imagery are then corrected by orthorectification using the Copernicus 90m Digital Elevation Model (DEM) called GLO-90. This step removes topographical distortions to give accurate spatial representation.
    
    \item \textbf{Normalization of Radar Backscatter:} The radar backscatter values are normalized using the terrain-corrected backscatter coefficient ($\gamma^0$), accounting for terrain effects. This leads to higher contrast between land and water pixels. In the 32-bit images, we get the actual radar backscatter, different from the quantized values given in the 8-bit images.
    
    \item \textbf{Upsampling:} Upsampling is implemented using bilinear interpolation to improve the image resolution, which degrades during the preprocessing. Upsampling preserves fine-grained details in coastline geometry, \eg~inlets and small bays, which are lost during preprocessing. Additionally, skip connections in U-Net tend to perform better with higher-resolution inputs.
\end{enumerate}

\subsection{Ground Truth Mask Generation}
Binary masks are required to represent land and water regions within each SAR image. These images are used as ground truth in our learning models. Manually generating ground truth masks for training and evaluation is time-consuming. Hence, we use an automated workflow to efficiently generate binary masks using land polygon shapefiles sourced from OpenStreetMap data~\cite{pelich2020coastline}. 

Our ground truth generation process has the following steps:
\begin{enumerate}
\item The spatial intersection between SAR images and land polygons is calculated.
\item The SAR images are processed to extract bounding coordinates, which are converted into bounding boxes and stored as GeoDataFrames. We used libraries, namely, rasterio and Shapely, for this step. The geocoding of SAR image boundaries is done using rasterio and these coordinates are converted into box-shaped polygons using Shapely, and stored in GeoDataFrames. This process creates a geographic ``template'' to clip out the relevant portion of a global land polygon dataset, which is then used to create binary masks where land is colored white (255) and water, black (0).
\item Land polygon shapefiles are overlaid onto these bounding boxes, creating intersection shapefiles that align with the SAR imagery.
\item These intersection shapefiles are used to generate binary land-water masks.
\end{enumerate}

\subsection{Coastline Classification Model}
We classify the SAR images into two categories: \textit{natural coastlines} and \textit{built coastlines}. They are similar to non-urbanized and urbanized coastlines~\cite{ciecholewski2024review}. Our proposed classification is based on perceptual differences, without any rigorous check on the level of \textit{urbanization} in those regions.

We hypothesize that the coastline extraction can be generalized for global datasets by using the additional information on the coastline type. Hence, we propose the use of coastline type-specific trained models for coastline extraction. This is not possible without automating the coastline classification itself. Hence, our integrated workflow, CCESAR, involves classification followed by segmentation for coastline extraction, as shown in Figure~\ref{fig:model}.

Our proposed classification model, $\cmod$, is based on a vanilla CNN architecture optimized for SAR imagery. It has an input layer followed by several convolutional blocks. Each block contains two convolutional layers with filter sizes progressively increasing from 32 to 256. Batch normalization and ReLU activation are applied to each convolutional layer to stabilize training and enhance feature extraction.

Max pooling layers are included to reduce the spatial dimensions of the feature maps, thus decreasing computational load while preserving critical features. A global average pooling layer condenses these feature maps into a single vector, which is passed through a fully connected dense layer of 512 units with ReLU activation. A dropout layer is applied for regularization, reducing the risk of overfitting. The final layer uses a sigmoid activation function, outputting a binary classification of natural or built coastline.

\subsection{Coastline Extraction Model}
For pixel-wise segmentation, the vanilla U-Net architecture is employed due to its encoder-decoder structure, which effectively preserves spatial information. The U-Net model has been used effectively for coastline extraction~\cite{ciecholewski2024review}. The U-Net consists of a contracting path (encoder) that captures context and an expansive path (decoder) that enables precise localization. Skip connections between corresponding layers in the encoder and decoder allow for the retention of spatial features at different resolutions, which is crucial for accurate coastline segmentation in SAR images. We train separate coastline extraction for different classes, thus, having $\sntr$ and $\sblt$ for natural and built coastlines, respectively.

\subsection{Model Training and Optimization}
The CNN and U-Net models were trained using the Adam optimizer with an initial learning rate of $1 \times 10^{-5}$. L2 weight decay regularization with a coefficient of 0.001 was applied to prevent overfitting. Both models were trained over 25 epochs with a batch size of 12, using binary cross-entropy as the loss function. The training was performed on a workstation equipped with an NVIDIA A4000 GPU to expedite the process, given the computationally intensive nature of the U-Net and CNN models.

\subsection{Post-Processing and Model Evaluation}
We then performed a post-processing step to extract the coastline from the segmented image. We used the widely used Canny edge detection algorithm to extract coastline boundaries from both the predicted and ground truth segmentation masks~\cite{ciecholewski2024review}. Hysteresis thresholding is the final step in the algorithm that uses lower ($\tau_l$) and upper ($\tau_u$) thresholds to preserve true edges and identify strong edges, respectively. In practice, a ratio of 1:3 is used as a rule of thumb for $\tau_l$:$\tau_u$. In our work, we found 50 and 150 to work well for coastline SAR images, respectively. Also, a standard Sobel filter of 3$\times$3 is found to be sufficient for gradient computation for applying the algorithm in this case. We finally select the longest contiguous edge detected in each mask as the final predicted coastline. 

We compute the average minimum Euclidean distance between the predicted and ground truth edge pixels to measure the spatial discrepancy between the predicted and actual coastline boundaries.

For the classification model, we compute the accuracy as the ratio of correctly classified images (natural or built coastlines) to the total number of images, providing an overall measure of the model performance.

For the segmentation models, we use the Intersection over Union (IoU) score as the performance metric. IoU is computed as the ratio of the intersection of the predicted segmentation and the ground truth segmentation to the union of the predicted and ground truth areas. This score provides a quantitative measure of overlap between the predicted and actual segmented regions, with a higher IoU indicating better performance. As an ablation study, the two models are tested across both classes of images to assess whether CCESAR improves segmentation accuracy compared to using a single U-Net model without classification. 

\section{Experiments, Results, and Discussion}
We tested CCESAR on our collated dataset described in Table~\ref{tab:datasets} and compared it with alternative models.

\begin{figure*}[t]
  \centering
  \fbox{\includegraphics[width=0.4\linewidth]{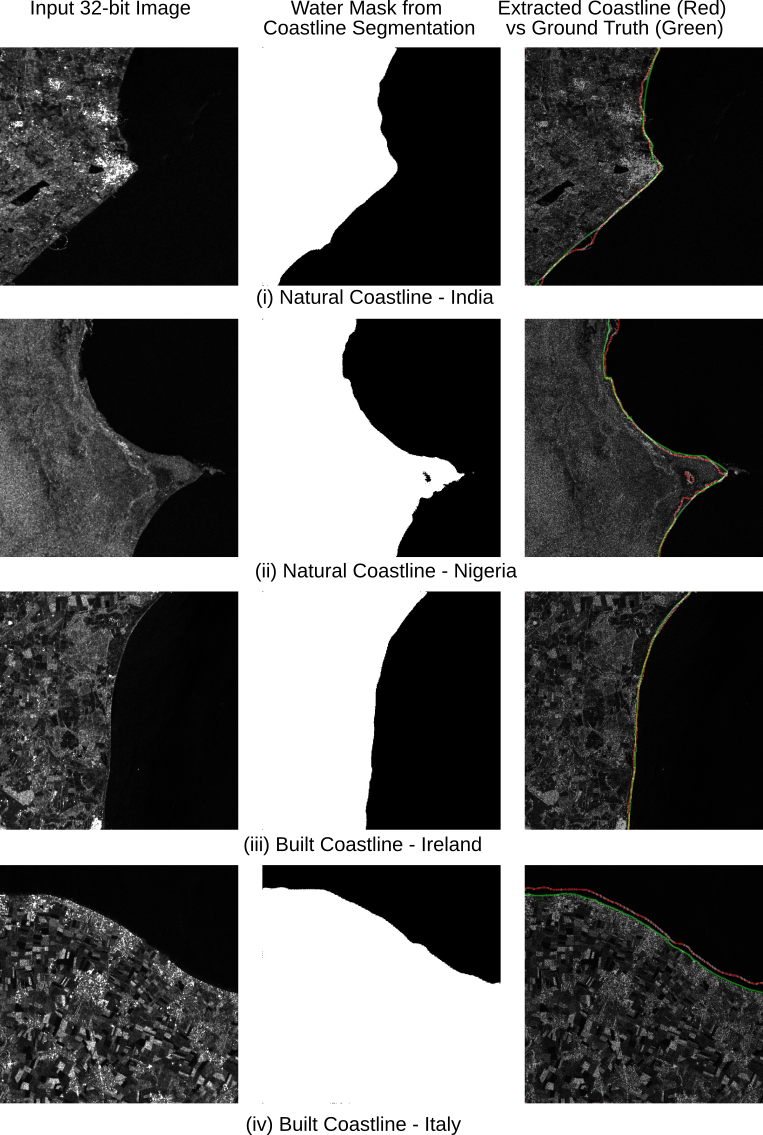}}
  \fbox{\includegraphics[width=0.4\linewidth]{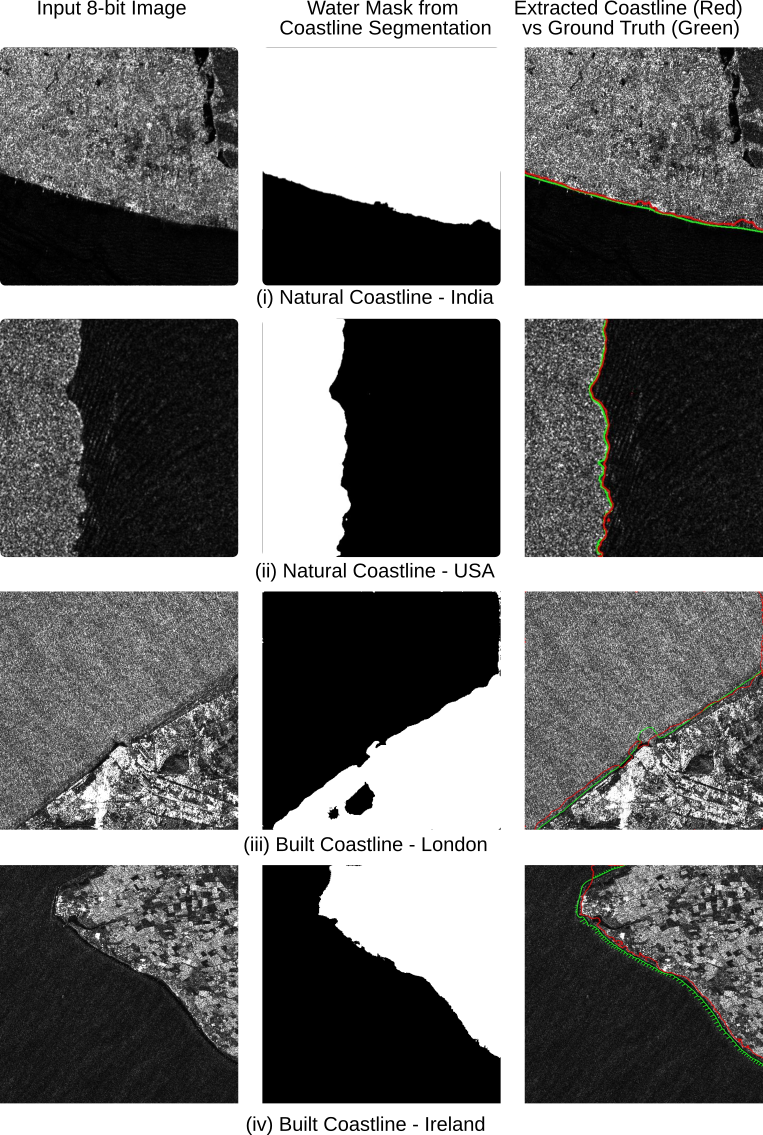}}\\
  \caption{Results from implementing CCESAR on both natural and built coastlines with coastlines extracted from Sentinel-1 SAR images with (left) 32-bit and (right) 8-bit compression.}
  \label{fig:results}
\end{figure*}

\begin{figure*}
  \centering
  \includegraphics[width=0.8\linewidth]{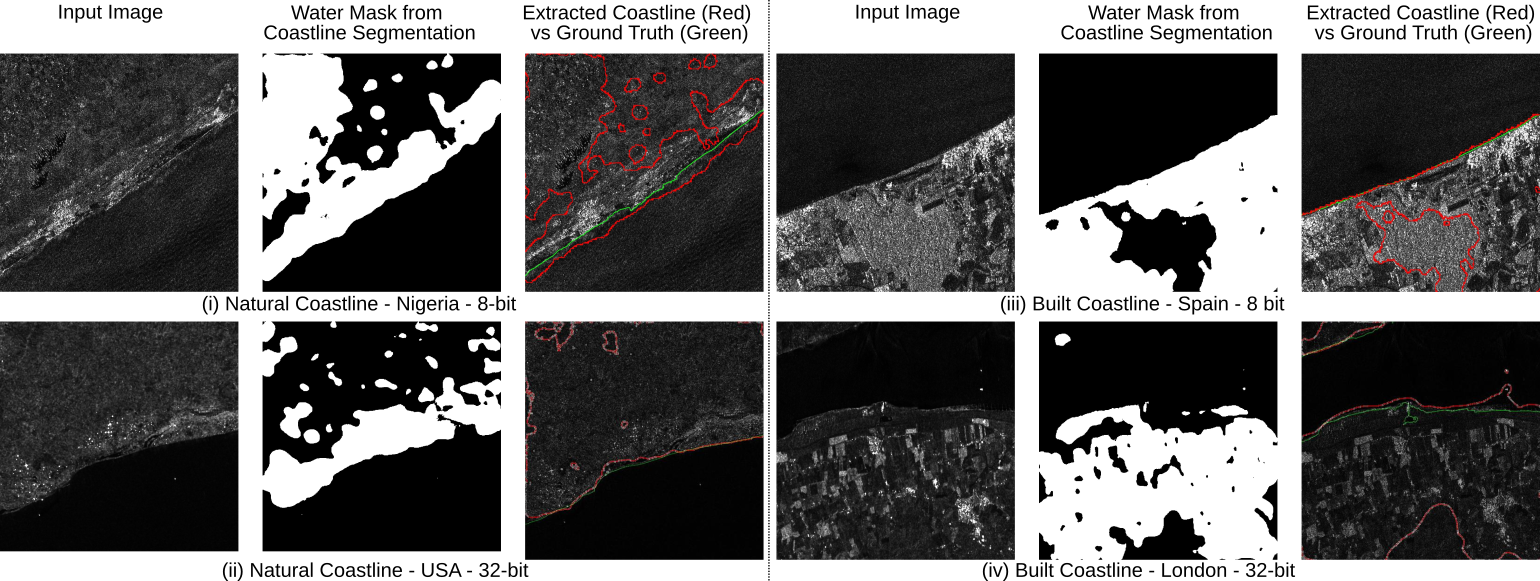}\\
  \caption{Examples of Sentinel-1 SAR images where CCESAR gives faulty results.}
  \label{fig:failures}
\end{figure*}

\begin{table}[htbp]
\centering
\caption{Performance Metrics for Different Experiments}
\label{tab:expt-results}
\scriptsize 
\begin{tabular}{|c|l|c|c|}
\hline
&& \multicolumn{2}{c|}{\textbf{IoU} (in \%-age)} \\\cline{3-4}
\textbf{E\#} & \textbf{Experiment Description}
& \textbf{8-bit Images} & \textbf{32-bit Images} \\ \hline
\rule{0pt}{3ex}    
\expt{1}& A Single Model on Both Classes & 41.63 & 59.27 \\ \hline
\rule{0pt}{3ex}   
\expt{2}& $\sntr$ on Unseen Natural Coastlines  & \textbf{96.17} & 93.99 \\ \cline{2-4}
\rule{0pt}{3ex}    
& $\sblt$ on Unseen Built Coastlines  & \textbf{95.04} & 91.56 \\ \hline
\rule{0pt}{3ex}    
\expt{3}& $\sntr$ on Built Coastlines  & 92.43 & 93.85 \\ \cline{2-4}
\rule{0pt}{3ex}    
& $\sblt$ on Natural Coastlines  & 49.91 & \textbf{96.30} \\ \hline
\rule{0pt}{3ex}    
\expt{4}& $\sntr$ on Both Classes  & 94.30 & 93.92 \\ \cline{2-4}
\rule{0pt}{3ex}    
& $\sblt$ on Both Classes  & 72.48 & 93.93 \\ \hline
\rule{0pt}{3ex}    
\expt{5}& CCESAR ($\cmod$ followed by $\sntr$ or $\sblt$)  & 86.59 & \textbf{95.15} \\ \hline
\end{tabular}

\vspace{1.5em}
\caption{Accuracy Results for 8- and 32-bit SAR Images}
\label{tab:accuracy}
\scriptsize
\begin{tabular}{|c|c|c||c|c|}
  \hline
  & \multicolumn{2}{c||}{\bf Classification} 
  & \multicolumn{2}{c|}{\bf Average Spatial}
  \\
  & \multicolumn{2}{c||}{\textbf{Accuracy}} 
  & \multicolumn{2}{c|}{\textbf{Discrepancy}} 
  \\\cline{2-5}
  \textbf{Image Class}
  & \textbf{8-bit} & \textbf{32-bit}
  & \textbf{8-bit} & \textbf{32-bit}
  \\ \hline

  Natural Images
  & 67.5\% & 87.5\%
  & 3.50 km & 3.31 km \\ \hline
  Built Images
  & 82.5\% & \bf 97.5\%
  & 4.00 km & 4.11 km \\ \hline
  Overall
  & 75.0\% & \bf 92.0\%
  & 3.75 km & 3.71 km \\ \hline
\end{tabular}

\vspace{0.25em}
\textit{The spatial discrepancy is measured in pixels, where 1 pixel is $\sim$0.01 km.}
\end{table}

\subsection{Experimental Setup}
A total of 40 unseen images from both natural and built coastlines were used for segmentation experiments, along with 40 additional natural images for generalized segmentation. The experiments were structured to compare the performance of the following experimental models, including ablation studies. It must be noted that in \expt{2} and \expt{3} described below, the classification model is not explicitly run during testing, instead the segmentation model is run on manually selected image sets. The descriptions of the experiments are as follows:
\begin{enumerate}
\item \expt{1}: Testing \textbf{a Single U-Net Model} -- We test a vanilla U-Net model trained on coastlines of both natural and built classes on unseen data of both classes, as an ablation study.
\item \expt{2}: Testing \textbf{the Trained Segmentation Model on Unseen Data of Intended Class} -- We test $\sntr$ and $\sblt$ on unseen data of the natural and built coastline images, respectively, to test the generalizability of the model within the intended class. 
\item \expt{3}: Testing \textbf{the Trained Segmentation Model on Unseen Data of the Other Class} -- We test $\sntr$ and $\sblt$ on data of the built and natural coastline images, respectively, to test the generalizability of the model outside of the intended class. 
\item \expt{4}: Testing \textbf{the Trained Segmentation Model on Unseen Data of Both Classes} -- We test $\sntr$ and $\sblt$ on data of both classes, unseen with respect to the training of all models in CCESAR workflow. This is to test the generalizability of the models across classes, without running the classification model.
\item \expt{5}: Testing \textbf{CCESAR Workflow on Unseen Data} -- We test the performance of the CCESAR workflow on unseen data with respect to the training of all models in the workflow. 
\end{enumerate}

\subsection{Results}
Our qualitative results are in Figures~\ref{fig:results} and~\ref{fig:failures}. We observe in Table~\ref{tab:expt-results} that, as expected, CCESAR workflow performance is high for 32-bit images, except for an outlier in the performance of \expt{3} for $\sblt$ on unseen natural images. The classification accuracy is low for 8-bit images (Table~\ref{tab:accuracy}), which explains the superior performance of \expt{2} of testing appropriate segmentation model for the corresponding class of images, \ie~manually classified images. Thus, for both 8- and 32-bit images, our work shows that classification before segmentation yields better results than using a single model, \ie~\expt{1}. We also observe that the spatial discrepancy of the extracted coastline is relatively low (Table~\ref{tab:accuracy}, as also can be qualitatively observed in the visualizations in Figure~\ref{fig:results}. CCESAR particularly fails when the land-water pixels are not perceptually separated by a line of contrasting pixels, or in the presence of water-like pixels within land (Figure~\ref{fig:failures}).

\mysubheading{Discussion}
\textit{Can a unified model of classification and segmentation for coastline detection be used?} Since the tasks involved are image classification followed by pixel classification, the models operate on different spatial scales. Since none of the existing models work on different models simultaneously, a unified model is not feasible. Hence, a two-stage network model is currently the only plausible option.

An immediate enhancement of our work in the future is to improve the image classification model for 8-bit SAR images.

\section{Conclusions}
This work demonstrates that the image segmentation model performs best for coastline detection on SAR images of different compression values when trained based on coastline types. Hence, an image classification model before the segmentation improves the outcomes of coastline detection and subsequent extraction from SAR images. 

\section*{Acknowledgment}
The authors would like to thank Abhinav Verma and Avik Bhattacharya for their guidance in creating the dataset.



\bibliographystyle{unsrt}
\bibliography{papers_coastline}
\end{document}